\documentclass[preprint,12pt]{elsarticle}

\usepackage{amssymb}
\usepackage{amsmath}
\usepackage{booktabs}
\usepackage{multirow}
\usepackage{array}
\usepackage{graphicx}
\usepackage{caption}
\usepackage{xcolor}
\usepackage{fontawesome5}
\usepackage{hyperref}
\usepackage{url}
\newcommand{\doi}[1]{\href{https://doi.org/#1}{\texttt{doi:#1}}}
\definecolor{orcidlogocol}{HTML}{A6CE39}
\newcommand{\orcidicon}[1]{%
    \href{https://orcid.org/#1}{\raisebox{-0.2ex}{\textcolor{orcidlogocol}{\faOrcid}}}%
}
\newcommand{\AuthorWithORCID}[2]{\orcidicon{#2}\,#1}

\makeatletter
\def\ps@pprintTitle{%
  \let\@oddhead\@empty
  \let\@evenhead\@empty
  \let\@oddfoot\@empty
  \let\@evenfoot\@oddfoot}
\makeatother

\begin{document}

\begin{frontmatter}

\title{Mathematical Framework for Custom Reward Functions in Job Application Evaluation using Reinforcement Learning}

%% ----- AUTHORS -----
\author{\AuthorWithORCID{Shreyansh Jain}{0009-0000-1693-5829}\textsuperscript{a,\textdagger,*}}
\author{\AuthorWithORCID{Madhav Singhvi}{0009-0004-6272-5127}\textsuperscript{c,\textdagger,*}}
\author{Shreya Rahul Jain\textsuperscript{a,\textdagger}}
\author{Pranav S\textsuperscript{b,\textdagger}}
\author{Dishaa Lokesh\textsuperscript{b,\textdagger}}
\author{Naren Chittibabu\textsuperscript{b,\textdagger}}
\author{Akash Anandhan\textsuperscript{b,\textdagger}}

%% ----- AFFILIATIONS -----
\address{\textsuperscript{a}Department of Computer Science and Engineering, SRM Institute of Science and Technology, Ramapuram, Chennai, India}
\address{\textsuperscript{b}Department of Computer Science and Engineering, Sastra University, Thirumalaisamudram, Thanjavur, India}
\address{\textsuperscript{c}Halıcıoğlu Data Science Institute, University of California San Diego, San Diego, United States of America}

%% ----- COMPANY & INFO -----
\address{\textbf{O6AI LABS}}
\address{\faGithub\ \href{https://github.com/CatalystAGI/A-Mathematical-Framework-for-Custom-Reward-Functions-in-Job-Application-Evaluation-using-RL}{\textbf{Source Code}}}
\address{\vspace{0.5em}\href{https://ieeexplore.ieee.org/document/11325393}{\textbf{\textit{Published in IEEE Xplore}}}}

%% ----- FOOTNOTES -----
\fntext[]{\textsuperscript{\textdagger}Work done during an internship at O6AI LABS.\\
Emails: \{shreyansh.jain, madhav, shreya, pranav, dishaa.l, naren, akash.anandhan\}@o6ai.com}

\fntext[]{\textsuperscript{*} Core Contributors}

%% ----- ABSTRACT -----
% Kept only one instance of the Abstract
\begin{abstract}
Most of the traditional Applicant Tracking Systems
(ATS) depend on strict matching using keywords, where
candidates that are highly qualified are many times disqualified
because of minor semantic differences. In this article, the
two-stage process of developing a more comprehensive resume
assessment system based on small language model that is trained
with fewer than 600M parameters is introduced and fine-tuned
by using GRPO with a unique-designed reward function.The
initial stage is (SFT) Supervised Fine Tuning, which are use to
create a strong base model with the ability to perceive resumes
beyond superficial overlap of keywords. This SFT model is
further-optimized in the second step with Reinforced Learning
(RL) via GRPO with the help of multi-component based
rewarding, which will not be considered as a commission of
tokens matching.In the initial RL experiments, we found a severe
difficulty in the shape of reward hacking: overly aggressive
penalty terms resulted in unstable training dynamics and
prohibitively negative model behaviour. This was solved by trial
and error refinement of the reward, and careful training
hyperparameter tuning, which led to a stable and controlled
process of gentle polishing.
GRPO-refined model shows high real-life performance, as it
shows accuracy of 91\% on unseen data used for testing. It has a
high recall of 0.85 on the SELECTED class with a perfect
precision of 1.0, which highlights its high reliability to be used in
identifying qualified applicants. These findings demonstrate that
an appropriately structured two-step fine-tuning pipeline can
effectively be used to transfer a small language model into
human-like candidate evaluation, surpassing shortcoming of both
traditional ATS systems and unrefined uses of reinforcement
learning.
\end{abstract}

%% ----- KEYWORDS -----
\end{frontmatter}

% --- MAIN CONTENT STARTS HERE ---

\section{Introduction}

With automation of industries rapidly provided by the development of Artificial Intelligence, recruitment is one of the most urgent fields in terms of technological change. Since only one vacancy can receive thousands of applications, screening of resumes manually is no longer feasible, and the common use of Applicant Tracking Systems (ATS) has become a standard practice. Nevertheless, the vast majority of current ATS solutions have inherent shortcomings: they rely heavily on keyword-based filtering methods and are highly insensitive to factors such as contextual skill relevance, depth of professional experience, and the quality of educational background. This has led to situations where many highly qualified candidates are overlooked, resulting in unfair selection processes and significant opportunity costs for organizations.

To address these weaknesses, this study presents a smart candidate assessment system based on Small Language Models (SLMs). In contrast to large-scale language models, SLMs, typically ranging between 200 and 600 million parameters, are computationally efficient, more predictable, and better suited for recruitment processes that require fine-grained reasoning. Their reduced tendency toward hallucination and improved performance in low-data regimes make them particularly effective for agentic recruitment systems~\cite{belcak2025small}.

A major contribution of this work is the proposed two-stage training strategy. In the first stage, Supervised Fine-Tuning (SFT) is applied to train the SLM on recruitment-related corpora, including job descriptions, resumes, and systematically encoded skill indicators, enabling the model to better understand hiring requirements. The second stage introduces reinforcement learning using Group-Relative Policy Optimization (GRPO)~\cite{shao2024deepseekmath}, guided by a custom reward function that accounts for skill diversity, professional experience, and educational background in a manner aligned with human recruiter judgment. This approach enables a deeper and more human-centered candidate evaluation process beyond the rigid constraints of traditional ATS systems.

The key contributions of this paper include: the design of a resume evaluation pipeline that replaces strict keyword thresholding with AI-based relevance ranking; empirical validation of Small Language Models in a highly specialized recruitment domain through domain-specific fine-tuning; and the first reported application of GRPO in human resources technology to align model outputs with expert human assessments. Experimental results obtained after 337 reinforcement learning steps on a dataset of approximately 3,000 resumes demonstrate that the proposed framework offers a scalable, fair, and effective talent acquisition solution suitable for modern recruitment environments.

\section{Literature Survey}

Traditionally, recruiting technologies have represented a trade-off between the effectiveness of automated screening and the fine-grained judgment of human evaluators. This domain has long been dominated by Applicant Tracking Systems (ATS), which have received persistent criticism due to their reliance on strict keyword-based filtering. Prior studies have demonstrated that such systems may inadvertently discriminate against qualified candidates because of subtle semantic mismatches~\cite{vanesch2021job}, a limitation further supported by recent investigations into algorithmic bias in hiring processes~\cite{albaroudi2024comprehensive}. This deficiency has motivated continued research into more advanced machine learning techniques capable of capturing deeper semantic relationships between resumes and job descriptions.

Early efforts in this direction leveraged transformer-based architectures and neural embeddings, including BERT-style and GPT-style models, to improve linguistic representation and matching accuracy~\cite{chavan2024enhancing}. While these approaches significantly enhanced performance, they also introduced a trend toward increasingly larger models, raising concerns related to computational cost, scalability, and real-world feasibility. More recent findings, particularly those reported by Belcak et al.~\cite{belcak2025small}, suggest a paradigm shift toward Small Language Models (SLMs) as a compelling alternative. These studies demonstrate that domain-specific fine-tuning, rather than sheer model size, plays a critical role in achieving strong performance, challenging the assumption that larger models universally outperform smaller ones.

In parallel, Reinforcement Learning from Human Feedback (RLHF) has emerged as a powerful framework for aligning language models with human preferences, as introduced by Stiennon et al.~\cite{stiennon2020learning} and Ouyang et al.~\cite{ouyang2022training}. Building upon this foundation, more recent optimization techniques such as Group-Relative Policy Optimization (GRPO) proposed by Shao et al.~\cite{shao2024deepseekmath} have demonstrated improved efficiency and stability in alignment processes.

Despite these advancements, reinforcement learning-based optimization continues to face challenges, particularly in the form of reward hacking. In such scenarios, models exploit weaknesses in the reward signal rather than genuinely optimizing the intended objective. This phenomenon can result in undesirable behaviors, including systematically pessimistic or biased evaluation patterns, as discussed in recent studies~\cite{gao2023scaling,tarek2025reward}. These concerns highlight broader unresolved issues in model alignment, as outlined by Casper et al.~\cite{casper2023open}.

Although substantial progress has been made, several important research gaps remain. The application of modern reinforcement learning methods, particularly GRPO, to human resource technologies using Small Language Models has received limited attention. Furthermore, empirical investigations into reward hacking within recruitment-oriented assessment systems are largely absent from existing literature. Current solutions also fail to provide holistic candidate evaluations that effectively align model outputs with the multi-objective decision-making processes of human recruiters. This work addresses these gaps by presenting the first reported application of GRPO for fine-tuning an SLM for resume evaluation, offering a real-world case study of reward hacking challenges in this context, and proposing a refined multi-component reward function as a practical mitigation strategy.

\section{Proposed Methodology}

The proposed AI resume evaluation agent is trained in two phases. Supervised Fine-Tuning (SFT) is first applied to get a baseline idea of the task, and then Generative Reward Policy Optimization (GRPO) is applied to refine the reasoning of the model to match expert-heuristic reasoning. The data was artificially created so that there was an equal representation of approval and rejection classes. Resume and job description templates and logical rules were used to create candidate resumes and job descriptions programmatically in order to simulate realistic recruitment conditions, but eliminate privacy concerns in real resumes. The dataset, whilst artificial, was to be internal consistent (skills, experience and outcomes) to offer a testbed of a valid evaluation.

\subsection{Model Selection and Configuration}

In the case of the base model, we picked \texttt{unsloth/Qwen2-0.5B-Instruct-bnb-4bit} as it is efficient and the best performance on the baseline. We used 4-bit quantization using the Unsloth library and PEFT through LoRA (rank = 16, $\alpha$ = 32), which allows us to perform efficient adaptation without refining all the parameters. The stage of SFT (3,000 samples (90\% train, 10\% validation)) was provided in the format of prompts where the model is being asked to perform as an HR expert and provide a response in the form of a JSON object with a score and binary status (SELECTED or REJECTED).

There were two epochs of training at a learning rate of $2 \times 10^{-4}$ (linear scheduler) and adamw-8bit optimizer. The optimal batch size was 8 and the per-device batch size was 2 and the number of gradient accumulation was 4. This configuration offered effective, memory-conscious training and also guaranteed consistent gradient updates as well as avoiding overfitting.

\subsection{GRPO Refinement Phase}

The second step would improve the SFT-tuned model, which would involve improving the quality and logical consistency of the evaluations. This is done by optimizing the policy of the model over a hand-crafted, multi-component reward function that is intended to promote more human-like, fined-grained reasoning and positively discourage the act of reward hacking.

\subsubsection{Reward Formulation}

The center of the GRPO stage is a reward function as in Eq.~\ref{eq:reward} which gives a single holistic feedback signal in a weighted combination of four criteria. This interdisciplinary nature is the main tool to combat reward hacking since the model needs to meet many, even conflicting, goals in order to reach a high reward and cannot be able to rely on a single, easy measure.

\begin{equation}
\text{Reward} = \sum_{i=0}^{N} (W_i * S_i)
\label{eq:reward}
\end{equation}

Where: (1) $N$ = number of evaluation criteria (here, $N = 4$); (2) $W_i \in [0, 1]$ weight assigned to criterion, subject to $\sum_{i=0}^{N} W_i = 1$; (3) $S_i$ = score assigned to criterion determined by task-specific rules; (4) $i$ is the index of the criteria.

The reward formulation proposed is based on the principle of weighted linear combination, which is similar to artificial neural network feature activation aggregation by weighted summations. The evaluation criteria have a proportional contribution to the total reward, so that the contribution of each factor is not dominant without the weight being explicitly specified.

\paragraph{Final Reward Calculation.} The weights and score ranges shown in Table~\ref{tab:parameters} were determined through iterative empirical tuning to maximize model stability and alignment during GRPO training. Multiple configurations were evaluated, and the final values were selected based on the best balance of reward sensitivity, classification performance, and avoidance of reward hacking.

\begin{table}
\caption{Parameter Configuration}\label{tab:parameters}
\centering
\footnotesize
\begin{tabular}{|l|l|l|l|}
\hline
\textbf{Parameter} & \textbf{Base} & \textbf{Score Range} & \textbf{Scoring Rules} \\
 & \textbf{Weight ($W_i$)} & & \\
\hline
Status Correctness & 0.40 & $S_i \in [-2, 2]$ & +2: TP; 0: TN; -1: FP; -2: FN \\
\hline
Score Accuracy & 0.20 & $S_i \in [-1, 1]$ & +1: Matches expected score; 0: \\
& & & Consistent with score; -1: \\
& & & Invalid/illogical \\
\hline
Skills Matching & 0.20 & $S_i \in [-1, 1]$ & +1: $\geq$75\% skill match; 0: 40--\\
& & & 74\% match; -1: <40\% or no data \\
\hline
Experience & 0.20 & $S_i \in [-1, 1]$ & +1: Score \& status match level; \\
Evaluation & & & 0: Partial alignment; -1: \\
& & & Misaligned \\
\hline
\end{tabular}
\end{table}

Early experiments showed that overly aggressive penalty ranges caused pessimistic model behavior, whereas more moderate configurations led to stable and human-aligned policy updates. The chosen formulation reflects the most stable configuration observed during tuning.

\begin{equation}
R = \sum_{i=1}^{4} (W_i * S_i)
\label{eq:final_reward}
\end{equation}

\subsection{Training Setup Notes}

GRPO training setup is carefully planned not just to be policy-optimal, but also to provide a solid defense against reward hacking, the behavior where a model uses the reward function to take advantage of the policy to score highly with nonsensical or undesirable outputs. The same 3,000 samples are used in this phase as in the SFT phase. This is a standard and intentional procedure, the idea of GRPO is not to learn anything new based on the labels of the dataset, but to optimize the reasoning policy of the model.

The multi-faceted reward function itself is our primary preventative tool, however, the training dynamics are the second level of defense which is critical. It starts with loading a SFT adapter, restoring the policy with a task-conscious, stable baseline. The set-up is then adjusted to a softer polishing instead of hard optimization. Very small learning rate of $2 \times 10^{-6}$ is used to make small and consistent policy changes.

It is important to note, and in contrast to the SFT phase, the GRPO training loop does not use an evaluation dataset deliberately. The refinement in the policy is solely informed by the reward signal produced by the training samples since the conventional measures of validation, such as accuracy, are not very useful in this optimization scenario. The model is trained only one epoch, or 337 training steps on our data. This short time limit is an intended option to restrict its exposure to the reward landscape and decrease the risk of over-optimizing.

The simplest system to directly overcome this policy drift is the application of KL-divergence regularization, whose regulation is determined by the beta parameter, which should be 0.1. This regularization punishes the model when the output policy of the model becomes too different to the original SFT policy. This functionally restrains the model to a space of realistic and sensible solutions that it is trained on during SFT and discourages it to produce bizarre and high-reward outputs.

\section{Results}

The experimental analysis shows that the model performance is dramatically improved throughout the two-phase training process as both the Supervised Fine-Tuning (SFT) and Generative Reward Policy Optimization (GRPO) stages provide positive results.

\begin{figure}
\centering
\includegraphics[width=\textwidth]{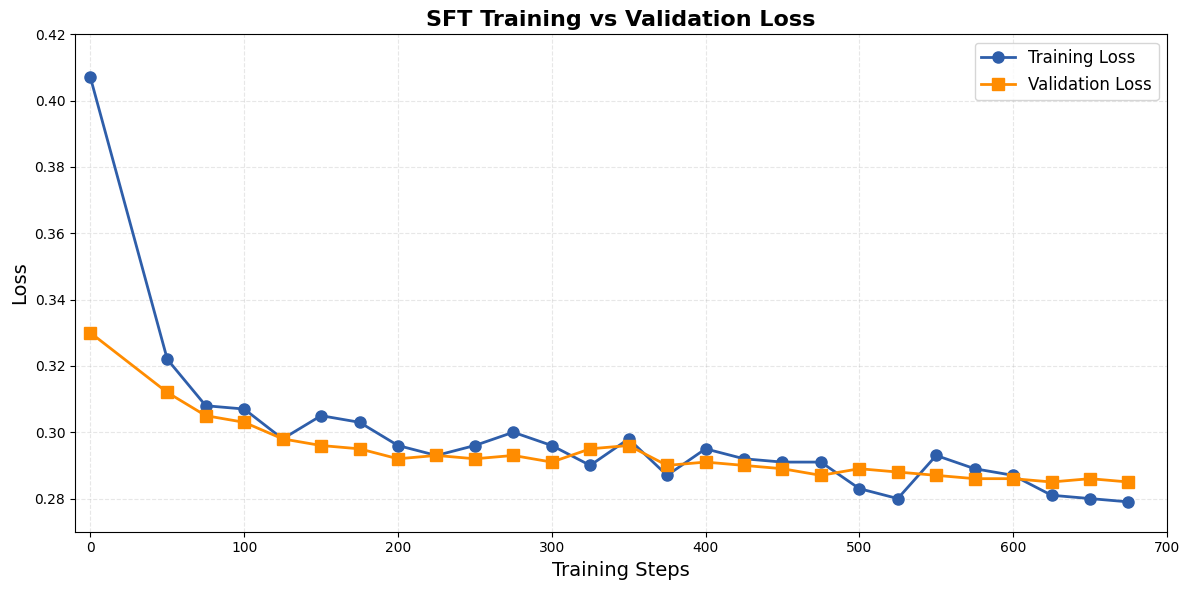}
\caption{Training and Validation Loss Curves for SFT} \label{fig:sft_loss}
\end{figure}

The SFT phase, which was done in two epochs on 2,700 training samples in total, managed to achieve a powerful baseline model. This training used a useful batch size of 8 and fine-tuned 8,798,208 LoRA parameters, 1.75\% of the total 502,830,976 parameters of the model. The training and validation losses to each other converged successfully as shown in Figure~\ref{fig:sft_loss}. The training loss reduced very fast initially, and the final value was around 0.28, which means that model was able to learn basic structure and format of resume evaluation task. Validation loss tended to follow training loss, as it reached a similar value, which proves that the model was not overfitting and was able to predict on the data that it had not seen before.

A GRPO optimization, which was performed over one epoch and took 337 steps, had more interesting and significant dynamics. The training loss reduced drastically by 97.3 percent (the initial training loss was 4.9380 and now it is 0.1337). This drastic decrease is an indication of the utility of the custom reward feature in the optimization of the policy of the model. This is also supported by the reward metric that evened out to a final outlook of $-0.0330$ meaning that the assessments of the model became more corresponding to our preferred requirements. At the same time, the KL divergence came to a final value of 0.34767, which demonstrates that the policy was optimized and stabilized without losing the knowledge acquired in the course of the SFT stage. Table~\ref{tab:comparison} provides a summary of these important metrics.

\begin{table}
\caption{Comparing Pre-trained and Post-trained Model}\label{tab:comparison}
\centering
\begin{tabular}{|l|c|c|c|c|c|}
\hline
\textbf{Method} & \textbf{Initial} & \textbf{Final} & \textbf{Loss} & \textbf{Final} & \textbf{KL} \\
 & \textbf{Loss} & \textbf{Loss} & \textbf{Reduction} & \textbf{Reward} & \textbf{Divergence} \\
\hline
SFT & 0.4070 & 0.2796 & 31.3\% & -- & -- \\
\hline
GRPO & 4.9380 & 0.1337 & 97.3\% & $-0.0330$ & 0.34767 \\
\hline
\end{tabular}
\end{table}

\begin{figure}
\centering
\includegraphics[width=\textwidth]{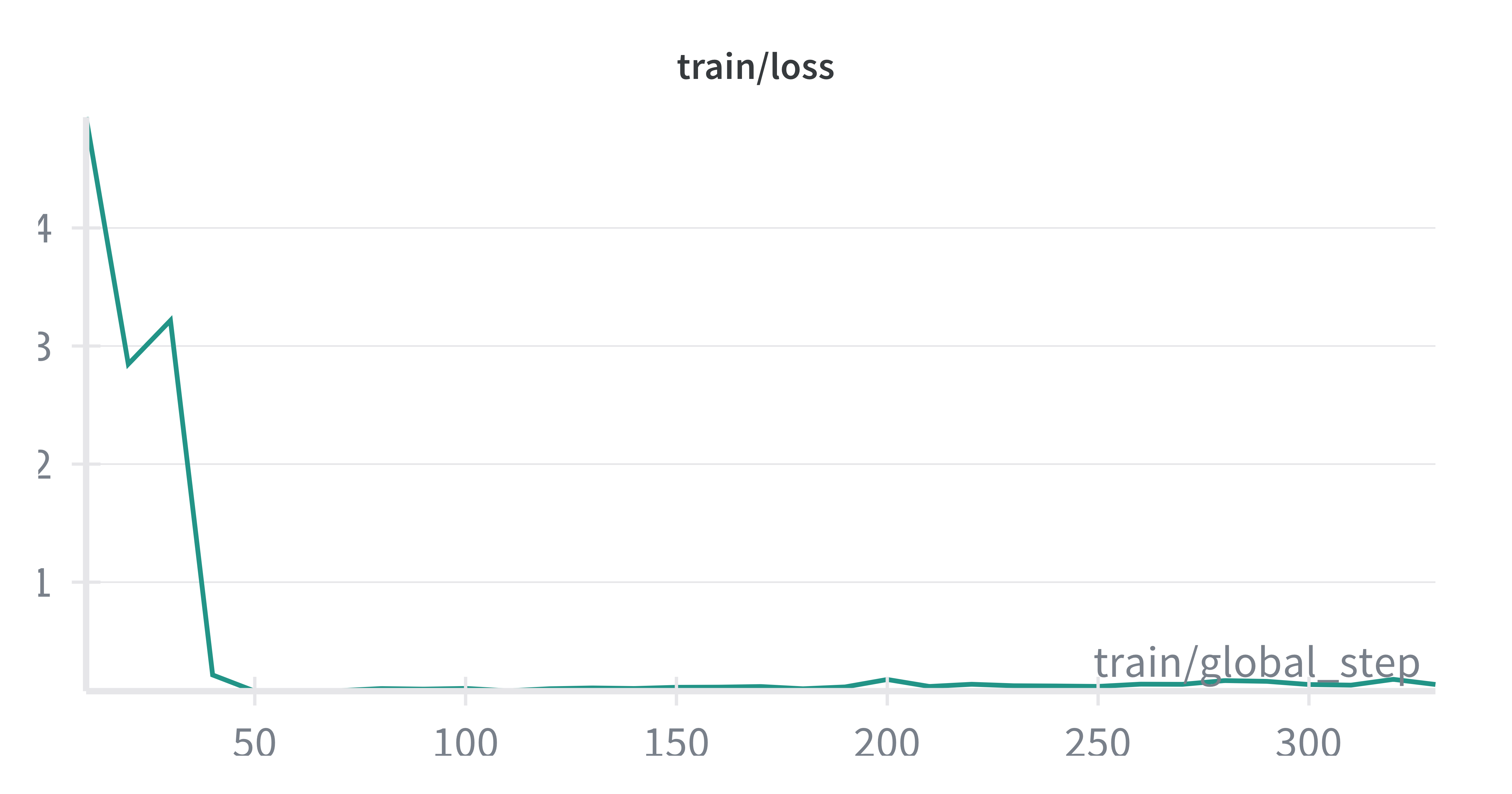}
\caption{GRPO Training Loss over 337 steps} \label{fig:grpo_loss}
\end{figure}

KL divergence metric is used to evaluate the extent of divergence of the policy of the model as compared to that of the original SFT policy. The trend has been equivalent to the training loss with an initial steep decline and then leveled to the low value as shown in Figure~\ref{fig:grpo_loss}. Such is the optimal behavior: it demonstrates that the model is making serious, constructive changes to its policy very early (large initial KL) but soon adopts a sophisticated state, without wandering too far out of the original knowledge it gained in the process of SFT (small final KL of 0.34767). This proves the fact that the training was balanced, and the so-called leash that the KL penalty offers worked in averting the collapse of the policy.

Subsequent examination of the training logs will show the consistency of the generation process. During the GRPO phase, the average length of the responses generated was always maintained between 60 and 90 tokens, and the average length of terminated responses was 30 to 50 tokens. This shows that the model had been trained to generate outputs of the appropriate and constant length. Moreover, the clipped ratio, indicating the share of policy changes, was changing, but tended to be in the 0.1 to 0.2 interval, indicating consistent and restrained policy changes during the course of training.

\subsection{Comparative Performance on Test Data}

To quantify the actual effects of the GRPO phase, both SFT-only model and the final GRPO-refined model were tested on the held-out test set of 104 unseen samples. GRPO-refined model performed better in all the key metrics compared to the SFT-only baseline. According to Table~\ref{tab:performance}, the GRPO model had a better classification accuracy (91.4\% vs. 89.4\%), and better F1-Score on the important class, i.e. selected (0.92 vs. 0.90). This validates that the phase of policy alignment did not only enhance the internal logic within the model but also carried over to the more precise final decisions.

\begin{table}
\caption{Comparative Performance on the Test Set}\label{tab:performance}
\centering
\begin{tabular}{|l|c|c|r|}
\hline
\textbf{Metric} & \textbf{SFT-Only} & \textbf{GRPO-Refined} & \textbf{Improvement} \\
 & \textbf{Model} & \textbf{Model} & \\
\hline
Overall Accuracy & 89.4\% & 91.4\% & +2.0\% \\
\hline
F1-Score & 0.9043 & 0.9204 & +1.8\% \\
\hline
Mean Absolute Error (MAE) & 16.05 & 15.47 & $-3.6\%$ \\
\hline
RMSE & 19.81 & 19.49 & $-1.6\%$ \\
\hline
\end{tabular}
\end{table}

\begin{figure}
\centering
\includegraphics[width=0.8\textwidth]{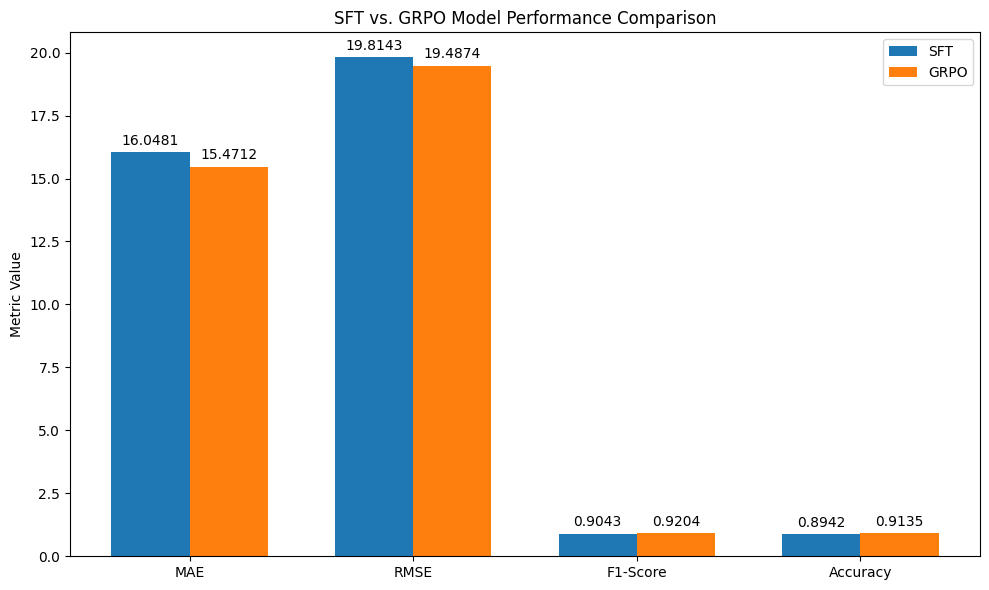}
\caption{Bar chart comparing SFT-only and GRPO-refined model performance} \label{fig:performance_chart}
\end{figure}

Figure~\ref{fig:performance_chart} provides a visual comparison of the performance metrics between the SFT-only and GRPO-refined models.

\begin{figure}
\centering
\includegraphics[width=\textwidth]{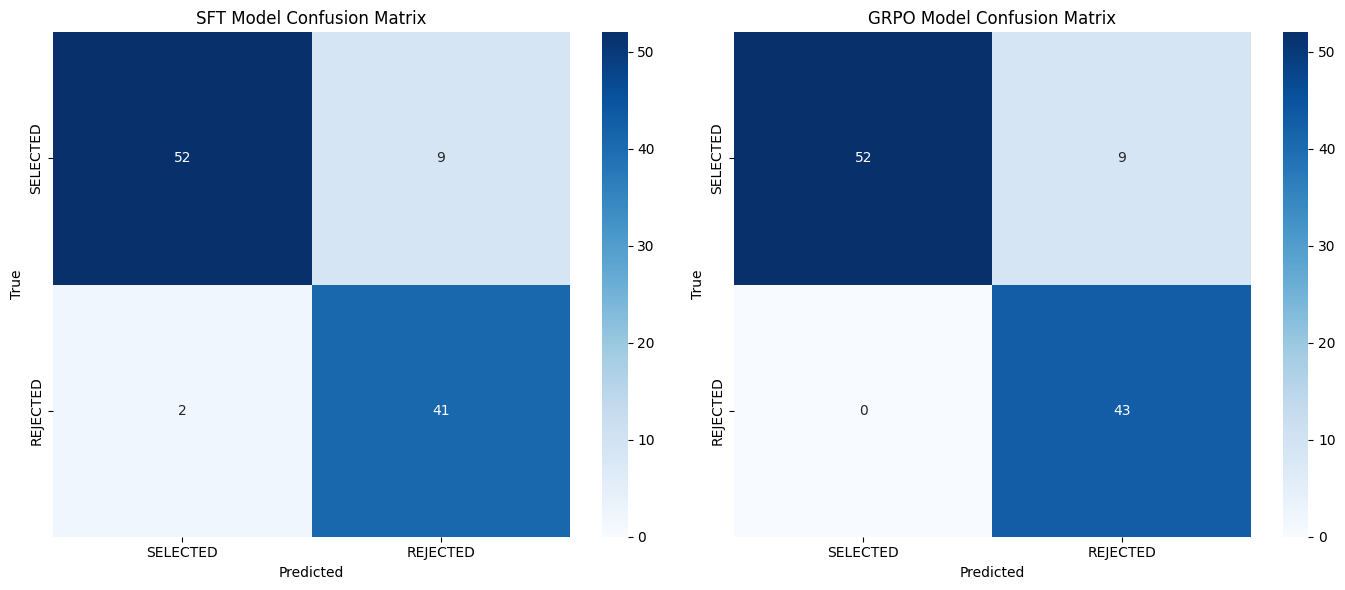}
\caption{Confusion matrices for SFT and GRPO models} \label{fig:confusion}
\end{figure}

When analyzed in granular detail through the confusion matrices shown in Figure~\ref{fig:confusion}, there was a crucial improvement of the decision-making of the GRPO model. The SFT model wrongly identified 2 REJECTED candidates as SELECTED. GRPO model removed these false positives altogether and the number of false positives dropped to zero, at the same time, the number of successfully identified `REJECTED' candidates (True Negatives) increased to 43.

This outcome is a direct success of the reward function's design. By penalizing incorrect classifications, the GRPO phase created a more discerning model that is less likely to pass unqualified candidates to the next stage, thereby improving the efficiency of the hiring pipeline.

\section{Conclusion}

This study manages to prove that a two-stage SFT and GRPO pipeline can be used to convert a small language model into an advanced resume judging system, despite traditional ATS being inflexible. Our last model gave 91 percent accuracy on unknown data and its ability to predict the chosen candidates was 98 percent, confirming its usefulness in the real world. The total success rate increase of the GRPO phase was 2.0 per cent, but the implications of these gains in practice are enormous. As an example, the refined model has zero false positive in the test set, the number of which was 2. In a practical hiring pipeline, this is a major time and cost savings as it means that unqualified applicants do not go through more resource-intensive steps.

Our main contribution is our multi-component reward function, which makes the model consistent with the complex business logic, most importantly, false negatives, and our training strategy of gentle polishing proved to be useful in reducing reward hacking. This paper introduces a computationally-efficient model building system to create expert models with nuanced and human-like reasoning, without the need to scale to massive architectures. The model can be further improved in the future to address its practical use further by allowing the model to identify candidacies of ambiguous grey areas and mark them as subject to manual review, to become a collaborative worker of decision-support.

\end{document}